\documentclass[letterpaper]{article} % DO NOT CHANGE THIS
\usepackage{aaai23}  % DO NOT CHANGE THIS
\usepackage{times}  % DO NOT CHANGE THIS
\usepackage{helvet}  % DO NOT CHANGE THIS
\usepackage{courier}  % DO NOT CHANGE THIS
\usepackage[hyphens]{url}  % DO NOT CHANGE THIS
\usepackage{graphicx} % DO NOT CHANGE THIS
\usepackage{natbib}  % DO NOT CHANGE THIS AND DO NOT ADD ANY OPTIONS TO IT
\usepackage{caption} % DO NOT CHANGE THIS AND DO NOT ADD ANY OPTIONS TO IT
\usepackage{amsmath}
\usepackage[dvipsnames]{xcolor}
\usepackage{multirow}

\usepackage{algorithm}
\usepackage{algorithmic}
\usepackage{subcaption}
\usepackage{cleveref}
\crefname{equation}{Eq.}{Eq.}
\crefname{section}{Section}{Sections}
\crefname{subsection}{Section}{Sections}
\crefname{subsubsection}{Section}{Sections}
\crefname{figure}{Figure}{Figures}
\crefname{table}{Table}{Tables}
\crefname{subfigure}{Figure}{Figures}
\crefname{algocf}{Algorithm}{Algorithms}

\newcommand{\eg}{e.g.}

\DeclareMathOperator*{\argmax}{arg\,max}

\usepackage{newfloat}
\usepackage{listings}
\usepackage{booktabs}
\DeclareCaptionStyle{ruled}{labelfont=normalfont,labelsep=colon,strut=off} % DO NOT CHANGE THIS
\floatstyle{ruled}
\newfloat{listing}{tb}{lst}{}
\floatname{listing}{Listing}
\title{
``Nothing Abnormal": 
Disambiguating Medical Reports 
\\
via Contrastive Knowledge Infusion
}

\author {
    Zexue He\textsuperscript{\rm 1},
    An Yan\textsuperscript{\rm 1},
    Amilcare Gentili\textsuperscript{\rm 1,\rm 2},
    Julian McAuley\textsuperscript{\rm 1},
    Chun-Nan Hsu\textsuperscript{\rm 1,\rm 2,\rm 3}
}
\affiliations {
    \textsuperscript{\rm 1} University of California, San Diego, La Jolla, CA\\
    \textsuperscript{\rm 2} VA San Diego Healthcare System, San Diego, CA\\
     \textsuperscript{\rm 3}VA National AI Institute, Washington, DC\\
    zehe@eng.ucsd.edu,
    ayan@eng.ucsd.edu,
    amilcare.gentili2@va.gov,\\
    jmcauley@eng.ucsd.edu,
    chunnan@ucsd.edu
}

\usepackage{bibentry}
\begin{document}

\maketitle

\begin{abstract}
Sharing medical reports is essential for patient-centered care.  A recent line of work has focused on
automatically generating reports with NLP methods. 
However, different audiences have different purposes when writing/reading medical reports -- for example, healthcare professionals care more about pathology, whereas patients are more concerned with the diagnosis (``\textit{Is there any abnormality?}''). The expectation gap results in a common situation  where patients find their medical reports to be ambiguous and therefore unsure about the next steps. In this work, we explore the \emph{audience expectation gap} in healthcare and summarize common ambiguities that lead patients to be confused about their diagnosis into three categories: \emph{medical jargon}, \emph{contradictory findings}, and \emph{misleading grammatical errors}. Based on our analysis, we define a disambiguation rewriting task to regenerate an input to be unambiguous while preserving information about the original content. We further propose a rewriting algorithm based on contrastive pretraining and perturbation-based rewriting. In addition, we create two datasets, OpenI-Annotated based on chest reports and VA-Annotated based on  general medical reports, with available binary labels for ambiguity and abnormality presence annotated by radiology specialists.  Experimental results on these datasets show that our proposed algorithm effectively rewrites input sentences in a less ambiguous way with high content fidelity. Our code and annotated data are released to facilitate future research.  
\end{abstract}

\section{Introduction}

Effective communication between healthcare providers and patients plays a critical role in patient outcome.
\emph{Patient-centered care}~\cite{catalyst2017patient,stewart2013patient} is reforming  traditional healthcare to shift a patient's role from an ``order taker'' to an active ``team member'' in their own healthcare process, to improve individual health outcomes and satisfaction~\cite{stewart2000impact}, and advocates sharing medical information fully and in a timely manner with patients.
It is also required by legal obligation (e.g.,~HIPAA\footnote{Shorten for Health Insurance Portability and Accountability Act} in the US) that patients have a legal right to access their personal health information. Failure of healthcare providers to communicate with patients efficiently and effectively about the results of
%the patients' 
medical examinations may lead to delays in proper treatment or malpractice lawsuits against providers~\cite{mityul2018patient,babu2015risk}.  

As a carrier of medical information, medical reports are shared with their patients by healthcare providers nowadays. 
% (e.g., Veterans Affairs\footnote{Shorten as VA. It is one of the largest healthcare systems in the world. VA shares imaging examination results including both reports and images with patients since 2018.} healthcare system of the US), as a part of the implementation of patient-centered healthcare. 
Medical reports serve many communication purposes with different audiences including ordering physicians, other care team staff members, patients and their families, and researchers~\cite{how2020reporting,gunn2013quality}. 
%They have 
Each group has
different needs and expectations
%from medical reports. 
%Usually the intended reader of medical reports is the other clinicians for internal communication instead of patients.
%When reading
when reading the reports:
%they have different needs and expectations --
peer medical professionals pay more attention to
actionable findings, while patients usually care more about the diagnostic outcome\footnote{We use exam result, diagnostic decision, abnormality existence  interchangeably, to express ``if there is anything abnormal''.} (i.e., \emph{Is there anything abnormal?}). How to address various communication needs for different audiences, 
%especially for patients 
and
to bridge the \emph{expectation gap between audiences} without increasing workload of report writers is critical.
\begin{table}[]
\resizebox{\columnwidth}{!}{%
\begin{tabular}{clc}
\toprule
 &
  \multicolumn{1}{c}{Report Sentence} &
  Diagnosis \\ \midrule
\begin{tabular}[c]{@{}c@{}}Medical \\ Jargon\end{tabular} &
  \begin{tabular}[c]{@{}l@{}}
  %\textbf{Am}: \textit{The right lung is \colorbox{pink}{grossly clear}.}
  {Am}: \textit{\colorbox{lightgray}{Unremarkable} bony structure.}
  \\ 
  %\textbf{Re}: \textit{The right lung is \colorbox{yellow}{normally clear}.}
   {Re}: \textit{\textbf{Normal} bony structure.}
  \end{tabular} 
  &
  Normal 
  \\ \cmidrule(l){1-3}
\begin{tabular}[c]{@{}c@{}}Contradictory \\ Findings\end{tabular} &
  \begin{tabular}[c]{@{}l@{}}{Am}: \textit{The lung volumes are \colorbox{lightgray}{low normal}.}\\ {Re}: \textit{The lung volumes are in the \textbf{lower}} \\ \textit{\textbf{half of the normal limit}.}\end{tabular} &
  Normal \\\cmidrule(l){1-3}
\begin{tabular}[c]{@{}c@{}}Misleading \\ Grammatical \\ Errors\end{tabular} &
  \begin{tabular}[c]{@{}l@{}}{Am}: \textit{Cardiomegaly and hiatal \colorbox{lightgray}{hernia}} \\ \textit{\colorbox{lightgray}{without} an acute abnormality} \\ \textit{identified}.\\ {Re}: \textit{Cardiomegaly and hiatal hernia \textbf{.} }\\ \textit{\textbf{Without} an acute ab-normality} \\ \textit{identified}.\end{tabular} &
  Abnormal \\ \bottomrule
\end{tabular}%
}
\caption{Ambiguous sentences (Am) from three categories with the unambiguous rewritten (Re). We highlight the parts causing ambiguity in gray, and show comparisons in bold.  }
\vspace{-1em}
\label{tab:ambiguous example}
\end{table}

To build such a bridge, it is important for medical reports to 1) be understandable with little specialized terms and 2) to have no ambiguity about the significance of findings when communicating with patients~ \cite{how2020reporting,mityul2018patient}. 
Previous works mainly focus on the first point where they change terminology to lay-person terms with replacement-based or deep learning methods~\citep{qenam2017text, oh2016porter, xu2022self}. %In real-world settings, the confusion caused by difficult special terminology (e.g.~rare pathology terms) usually arises from unreadability and can be alleviated by finding more information about the subject online. 
However, how to mitigate the ambiguity in a comprehensible report is crucial but rarely investigated.

% AI has much to offer to address this problem. Recent efforts have investigated using NLP models to assist healthcare practitioners, e.g.~by automatically generating accurate reports given images from a medical exam \cite{jing2018automatic, li2018hybrid, liu2019clinically, ni2020learning, yan2021weakly}. Potentially, AI report generators can be trained to generate reports that match the needs and expectation of patients.

In our work, 
we consider medical reports written in free text and analyze the ambiguity where patients are unsure about their exam results.
We first collect medical report data and ask domain specialists to label the binary abnormality presence associated with each sentence, and non-experts to label sentences that they deem ambiguous. Our medical team analyze the results and categorize the major causes behind ambiguity primarily into three categories: the report sentence is ambiguous due to containing (1) \textbf{medical jargon} with meanings different from everyday general usage, such as \textit{unremarkable};
%unremarkable}; 
(2) \textbf{contradictory findings} in the same sentence; 
(3) \textbf{misleading grammatical errors} such as no period between full sentences. Examples are shown in \cref{tab:ambiguous example}. 

%Other works that notice the communication gaps between medical professionals and patients usually focus on changing terminology to lay-person terms, using replacement-based methods based on dictionary or lexical rules\cite{qenam2017text, oh2016porter}, or leveraging style transfer with a deep model \cite{xu2022self}. It is worth mentioning that the confusion caused by difficult special terminology (e.g.~rare pathology terms) is beyond our scope, as it arises from the unreadability and can be alleviated by finding more information about the subject online. However, in this study, we highlight the  confusion coming from readable texts where a search may be insufficient. 

To alleviate patient confusion, we propose a new task called \emph{medical report disambiguation}, which is defined as: given an ambiguous sentence from a medical report that patients find hard to understand the 
%examine 
exam
result (``\textit{Is there any abnormality?}''), rewrite it with minimal edits in a way that the diagnostic decision is expressed to be more %tendentious
explicit, while retaining the precision of the original findings,
%health condition, 
namely preserving the detected pathologies and preventing new 
interpretations
% findings or diseases 
from being implied.

Paraphrasing models may offer a solution, but they are limited by the need of parallel corpus, which requires significant workload from radiologists.
% a heavy burden for specialists. 
To alleviate the annotation burden, we propose a rewriting framework without parallel corpora for disambiguation (see \cref{fig:model}). We first pretrain a Seq2Seq model in the medical domain with contrastive learning. Then, an ambiguous input is rewritten using the model by perturbing its hidden states and pushing the generation towards a direction that is more explicit about its exam results.
%pushing the re-generation  more explicit about its original exam results.
The pretraining step not only enables a model to capture the underlying language distribution for writing a human-readable medical report, but also enforces a property that sentences sharing similar pathology patterns will reside closely in the latent embedding space.
%which limits feasible perturbations still within niches. 
The two steps work together to preserve content fidelity with original input.

In summary, our work makes the following contributions: (1) We explore a novel and important problem in the healthcare domain regarding ambiguous medical reports. We empirically analyze the common reasons for ambiguous reports that make patients confused, and formally define the disambiguation rewriting task. (2) Based on our analysis, we propose an effective rewriting framework. Our model does not require parallel ambiguous and rewritten ``golden'' sentences for training, which alleviates the workload of medical specialists. (3) In addition, we provide two new datasets, OpenI-Annotated in chest radiology imaging and VA-Annotate data in general medical domain\footnote{Our annotated data and source codes will be released at  \url{https://github.com/ZexueHe/Med-DEPEN}} , each annotated with high-quality labels for ambiguity and abnormality presence from radiology specialists. (4) Using these datasets, we perform experiments and evaluate rewritten results based on disambiguation and fidelity preservation.
The results of both automatic and human evaluations indicate the effectiveness of our proposed method. 

To the best of our knowledge, our work is the first attempt to build an AI system to deal with patient confusion caused by ambiguous reports, which can potentially help promote patient-centered healthcare.

% \subsection{Contradicting Findings}
% \subsection{Misleading Grammatical Errors}
% \subsection{Medical Jargon}
\fboxrule=0.3pt%border thickness
\fboxsep=0mm
\section{Disambiguating Rewriting Framework}
\begin{figure*}
    \centering
    \subcaptionbox{Contrastive Pretraining}[.427\textwidth]{
    \centering
    \frame{\includegraphics[width=0.98\linewidth]{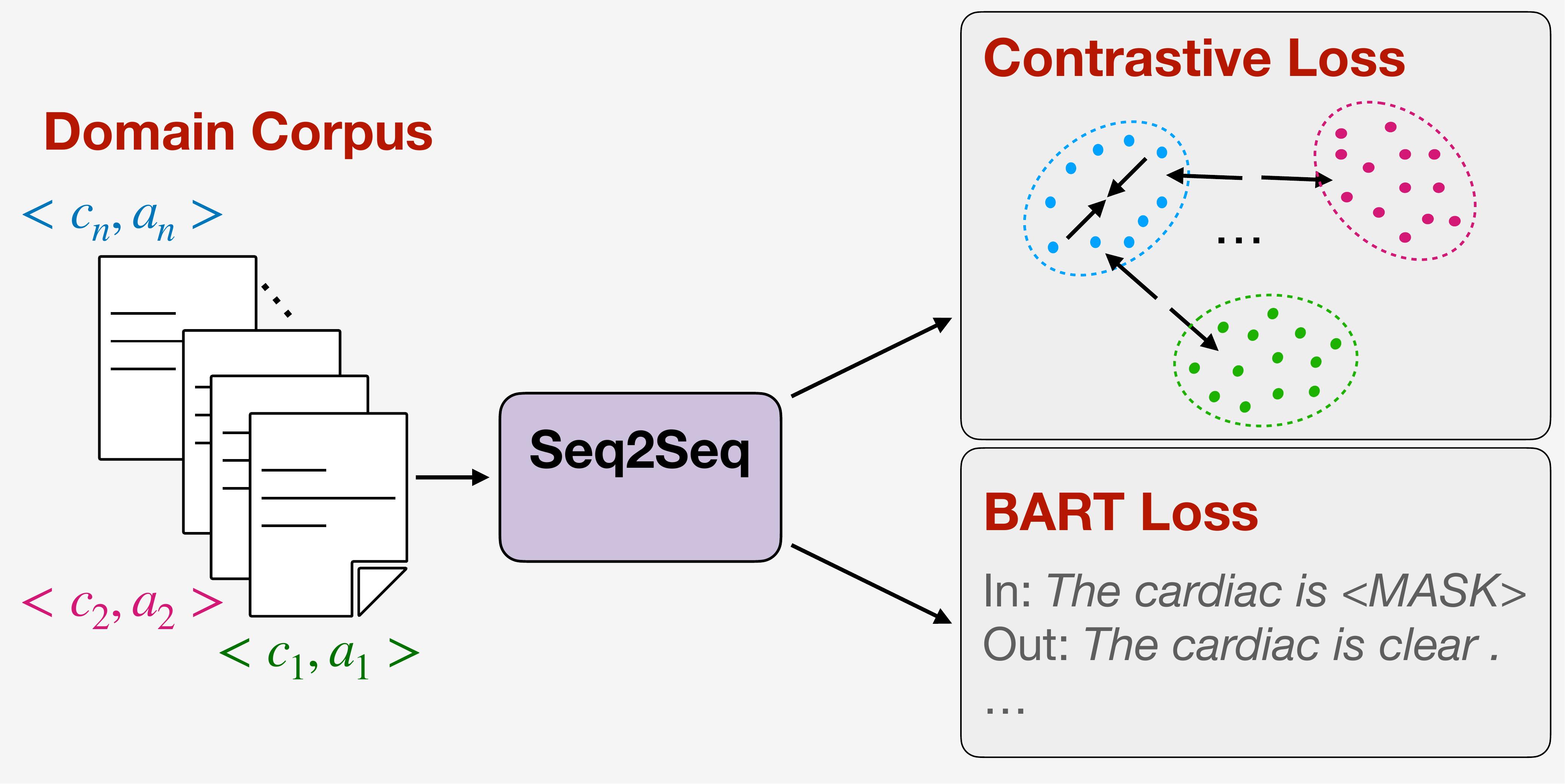}}
         \label{fig:model_pretrain}
         }
     \subcaptionbox{Rewriting Framework}[.552\textwidth]{
     \centering
          \frame{\includegraphics[width=0.98\linewidth]{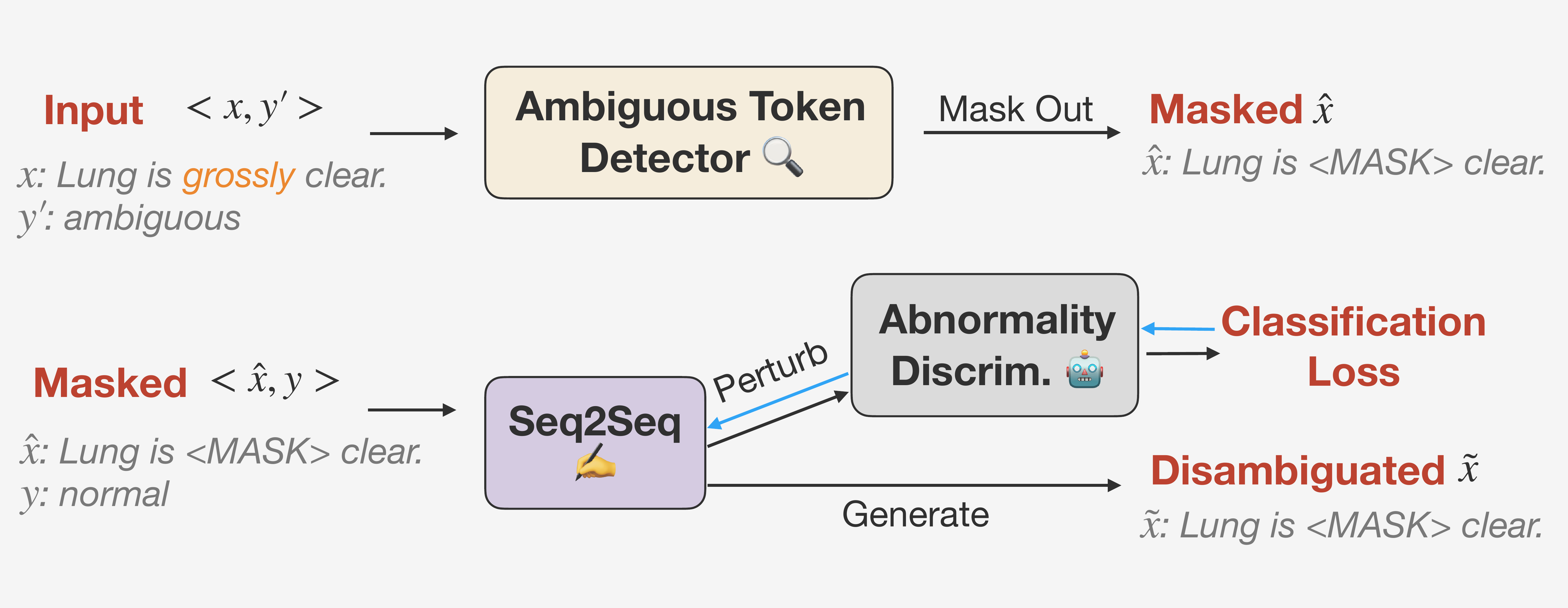}}
         \label{fig:model_rewrite}
     }
    \caption{Model illustration. Our model contains two steps: first do (a) constrastive pretraining  and then  (b) rewriting.}
    
    \label{fig:model}
\end{figure*}
Our task is to disambiguate an input medical sentence when a patient finds it  hard to understand the diagnostic decision. For an ambiguous sentence $x$ whose abnormality label is $y$ (abnormality presents
% exists 
or not), we will output a disambiguated sentence $\Tilde{x}$ that is more explicit about $y$.

We propose a contrastive knowledge infused rewriting framework to achieve this goal, which comprises a pretraining step and a rewriting step, as shown in \cref{fig:model} (a) and (b). We first obtain a medical-domain Seq2Seq model $\mathcal{G}$ that can effectively capture language patterns in different health situations in the pretraining step, and we generate a less ambiguous sentence using $\mathcal{G}$ in the rewriting step. We introduce each step in following sections. 

\subsection{Contrastive Pretraining}
First, we pretrain a domain-specific Seq2Seq model to generate medical language on top of a general domain BART~\cite{lewis2020bart}.
% \footnote{We call this procedure  \emph{pretraining}. }
%therefore, we pretrain a medical $\mathcal{G}$ with Mask Language Modeling task by optimizing the $\mathcal{L}_{MLM}$.
For our task, a pretrained model that only captures the distribution of medical language is not precise enough  -- there are several ways to rewrite an abnormal diagnostic in order to make it ``more abnormal''. 
%Take an example of 
Consider a patient with a diagnosis % symptom 
of having excessive
% abnormal 
lung fluid. This abnormal diagnosis can be rewritten to be ``more abnormal'' by combining it with other abnormalities
% symptoms 
(such as \emph{unusual liquid and unusual air}) or by changing the disease (from a lung disease to % \emph{another} lung 
a heart disease). 
This is undesirable.  
Therefore, we require the rewritting to 
% a certain amount of high precision about the original diagnosis, which means 
preserve the original diagnosis. %and prevent from introducing new one.
To achieve this goal, more domain knowledge about different pathologies is required.
%about the healthcare condition is required. 
%Due to the lack of fine-grained disease labels for rewriting input, 
We capture such domain knowledge by learning from external corpora (such as MIMIC-CXR\citep{johnson2019mimic}) that are on a large scale in the same medical domain with fine-grained disease labels. We pretrain the language model by infusing the domain knowledge with supervised contrastive learning, which pushes sentences closer if they express
% share 
similar pathological findings
% patterns 
and pull away sentences if they are not. As a result, we can reduce the probability of rewriting a sentence medically different from
% unfeasible sentence for 
the original input. This is crucial for patient safety and a unique issue in our task different from other  text rewriting problems.

%Here we introduce our contrastive pretraining strategy for transformer-based seq2seq model. 
The external medical corpora consist of medical report pairs including both sentence $c_i$ and its associated fine-grained pathological label $a_i$ (such as disease labels like \emph{atelectasis, edema, no finding, etc}). We pretrain an encoder-decoder transformer $\mathcal{G}$ with supervised contrastive learning, following \citet{khosla2020supervised}.  For each sentence $c_i$ in a mini-batch $B$, we first obtain its representation $H_{c_i}$ by taking the last hidden states  from the decoder in $\mathcal{G}$. Then a $\tau$-temperature similarity $s_{i,j}$ between $c_i$ and another sentence $c_j$ in $B$ is calculated:
\begin{align}
    s_{i,j}=\text{sim}(H_{c_i}, H_{c_j}) = H_{c_i} \cdot H_{c_j}/\tau
\end{align}

We use $S(i) = \{c_j: a_j=a_i\}$ to denote the set of sentences sharing the same disease label $a_i$, then the contrastive learning loss $\mathcal{L}_{\text{CL}}$ for the mini-batch $B$ is defined as  
\begin{align}
\mathcal{L}_{\text{CL}} &= \sum\limits_{c_i \in B}  \mathcal{L}_{\text{CL}, i} \end{align}
where 
\begin{align}
    \mathcal{L}_{\text{CL}, i}&=-\frac{1}{|S(i)|}\log \frac{\sum\limits_{c_j\in S(i)}\exp(s_{i,j})}{\sum\limits_{c_j \in B } \exp (s_{i,j}) }
    %\nonumber
    \label{eqn:contra}
\end{align}
% Here, $s_\{i,j\} = \text{sim}(z_i, z_j)$ is the similarity function of example $x_i$ and $x_j$. We use $ z_i \cdot z_j / \tau$ to meassure the similarity, where $z_i$ and $z_j$ is the last hidden representation of decoder for $x_i$ and $x_j$, and $\tau$ is the temperature. 

Our constrative pretraining is also applicable even when the pathological  label $a_i$ is not available. 
A recent empirical study \cite{oakden2020hidden} shows  that the  representations from deep neural networks carry information of labels and unlabeled features. 
Inspired by this, in the case where $a_i$ is not available, we first extract sentence representations from a medical Bert pretrained with a radiology report corpus \cite{yan2022radbert}. Then we follow  \citet{sohoni2020no} to cluster sentences with a Gaussian Mixture Model. The clustered results carry fine-grained information about different pathological patterns, and work as an approximation of the labels used in optimizing \cref{eqn:contra}.

Besides the contrastive learning objective, our pretraining also includes  a token  infilling task ~\citep{lewis2020bart} in order to obtain an informative representation $H_{c_i}$, which reconstructs the original sentence $c_i$ from its randomly masked version $\hat{c_i}$:
\begin{align}
    \mathcal{L}_{\text{BART}} = -\sum_i^{|B|}\sum_t^{|c_i|} \log p(c_i^t| c_i^1, \cdots, c_i^{t-1}; \hat{c_i})
\end{align}
Therefore, our pretraining goal is to learn the medical language distribution (language modeling loss $\mathcal{L}_{\text{BART}}$) and capture language patterns of different medical conditions (contrastive loss $\mathcal{L}_{\text{CL}}$), and is formulated by minimizing their weighted sum  in \cref{eq: pretrain obj}:
\begin{align}
    \mathcal{L}=\lambda_1\mathcal{L}_{\text{BART}} + \lambda_2 \mathcal{L}_{\text{CL}} \label{eq: pretrain obj}
\end{align}

\subsection{Rewriting Framework}
% \subsubsection{Mask Out Ambiguous Parts}
% \subsubsection{Rewrite for De-ambiguity}
During the rewriting process for an ambiguous input $x_i$, the following objectives are targeted: 1) the main content is retained, and 2) the diagnostic decision is more explicitly expressed
% reflected 
in the rewritten sentence. While contrastive pretraining ensures a reasonable level of content fidelity, the first objective also suggests minimal changes during rewriting, which only touch those portions necessary for disambiguation. The second one requires a controllable generation that pushes the generation closer to the diagnostic decision.

Inspired by recent advance in controlled text generation~\citep{he2021detect}, we leverage a plug-and-play method to rewrite the sentences without the need of parallel annotated training data. It includes a \emph{detect} stage to mask potential tokens that are highly predictable for an attribute and a \emph{perturb} stage to do neutralization rewriting w.r.t.~that attribute.
%gender. 
Since we need to detect tokens in $x_i$ that are highly predictable in their ambiguity, we first train an ambiguity classifier during detect stage. The tokens with the top-K highest attention scores will be detected as salient for ambiguity and will be masked. Then in the perturb stage, we require an edit that is more explicit in the direction of its diagnostic decision $y_i$, rather than making it more neutral for the ambiguity. Therefore, we modify the perturb stage to suggest an explicit edit by maximizing the likelihood of making the right diagnostic decision at each generation step $t$:
\begin{align}
    \Tilde{x}_i^t=\argmax_{\Tilde{x}_i^t} p(y|\Tilde{x}_i^t),
    \label{eq: purturb}
\end{align}
where the distribution $p$ is output from a classifier $f$ which predicts the diagnostic decision $y_i$, pretrained by minimizing $\text{Cross-Entropy}(f(x_i), y_i)$. 

Then, during generation, we add a perturbation to decoder hidden states in $\mathcal{G}$ by taking the gradient w.r.t.~$\Tilde{x}_i^t$ from the Cross-Entropy loss, 
and regenerate the token distribution
%for this time, 
since the hidden states have been updated. Alternatively, adding perturbation and (re-)generation push the rewritten sentence towards the direction of its 
%abnormality polar, 
diagnostic decision, 
that is to say, being less ambiguous.
% \begin{align}
%     \Delta H_t &\leftarrow \Delta H_t + \lambda \nabla f(y|H_{t}+\Delta H_t) \\
%     \Tilde{x}_{t+1} &= Dec(\Tilde x_t, H_{t}+\Delta H_t)
% \end{align}

\section{Experimental Setup}
%Here we discuss the experiment setups, which are organized as follows:
%First, we provide an overview of our experiment settings. 

Our rewriting algorithm is tested in two practical settings. First, disambiguating chest reports in a specialized medical domain. Secondly, disambiguating general medical reports that cover many imaging modalities (e.g., x-ray, CT, etc.) and body parts.
% organs and parts of the body. 
Our medical team created 
% annotated rewriting 
annotation datasets (OpenI-Annotated and VA-Annotated) for each experiment. During pretraining, an additional large-scale medical corpus is used in each experiment. 

%Then, we discuss the datasets used in our task, including two types: \textit{rewriting datasets} with human binary annotations for ambiguity and abnormal findings, and \textit{pretraining datasets} with fine-grained contrastive labels. Statistics and labeling procedures are also discussed. Finally, we introduce the baseline models and evaluation metrics

% \subsection{Experiment Setting Overview}
% Our rewriting algorithm is tested in two practical settings. First, disambiguate chest reports to test the precision of rewriting models in a narrow medical domain. Secondly, disambiguate general medical reports that cover many organs and parts of the body. In this experiment, we investigate the ability to rewrite models when the covered objects are messy and noisy. Our medical team creates annotated rewriting data (OpenI-Annotated and VA-Annotated) for each experiment. During pretraining, large-scale medical data is used in each experiment. 

\subsection{Human-Annotated  Datasets for Rewriting}
\label{sec: rewriting dataset}
\begin{figure}
    \centering
     \includegraphics[width=0.9\columnwidth]{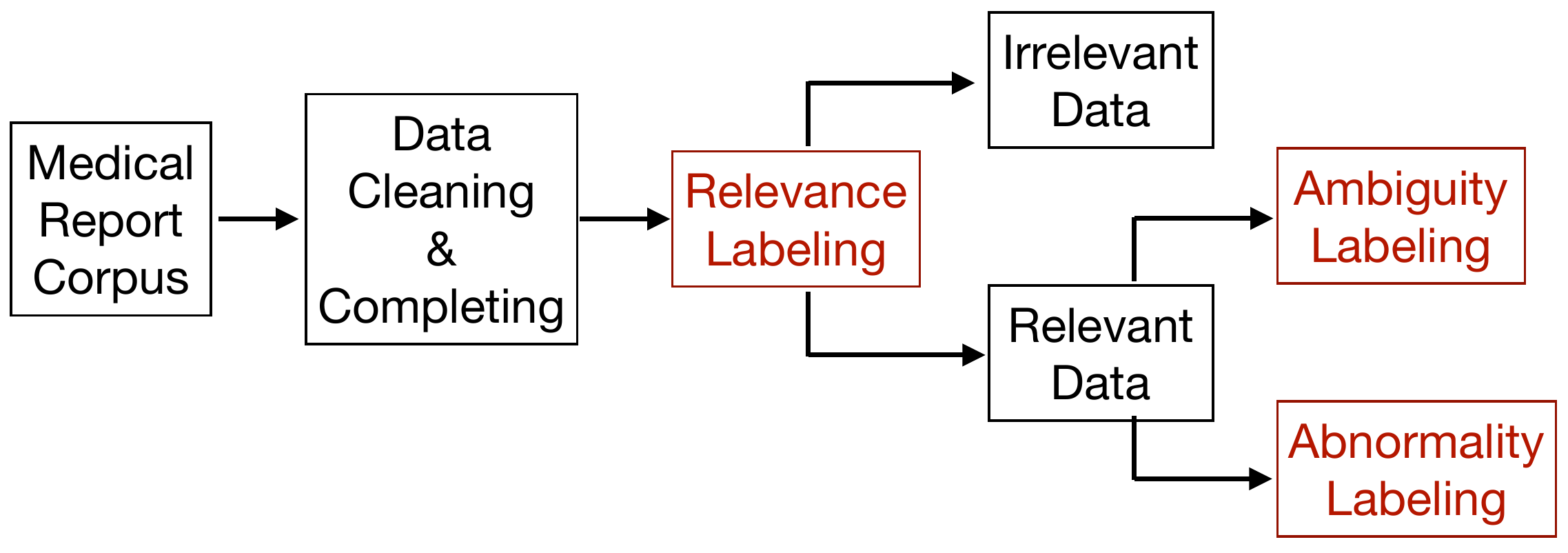}
    \caption{Data Annotation Pipeline. Three different labels are annotated in red steps.}
    \label{fig: data annotation}
\end{figure}
\begin{table}[]
\centering
\resizebox{0.9\columnwidth}{!}{%
\begin{tabular}{cccc}
\toprule
                                                    Dataset        & Total  & Ambiguous & Abnormal \\ \midrule
OpenI Annotated & 15,023 & 988       & 6,111    \\
VA Annotated    & 5,180  & 1,461     & 2,358    \\ \bottomrule
\end{tabular}%
}
\caption{Statistics of Annotated Datasets}
\vspace{-1em}
\label{tab:datasets}
\end{table}

\begin{table}[]
\resizebox{\columnwidth}{!}{%
\begin{tabular}{cccc}
\toprule
Disease       & Num.  & Disease & Num.   \\ \midrule
Enlarged Cardiomediastinum & 17,944 & Atelectasis          & 32,445  \\
Cardiomegaly               & 56,099 & Pneumothorax         & 5,539   \\
Lung Opacity               & 62,865 & Pleural Effusion     & 36,537  \\
Lung Lesion                & 9,838  & Pleural Other        & 3,350   \\
Edema                      & 14,605 & Fracture             & 10,893  \\
Consolidation              & 3,905  & Support Devices      & 8,355   \\
Pneumonia                  & 1,365  & No Finding           & 865,738 \\ \bottomrule
\end{tabular}%
}
\caption{Fine-grained diseases in MIMIC-CXR. }
\label{tab:mimic conditions}
\end{table}

% \usepackage{multirow}
% \usepackage{graphicx}
\begin{table*}[]
\centering
\resizebox{0.9\textwidth}{!}{%
\begin{tabular}{cccccc}
\toprule
\multirow{9}{*}{\begin{tabular}[c]{@{}c@{}}\textbf{OpenI-}\\ \textbf{Annotated}\end{tabular}} &  & \textbf{Ambiguity Acc.} & \textbf{Decision Acc.} & \textbf{Pathology Match} & \textbf{PLL} \\ 
 &
  \textbf{Raw Text} &
  0.855 &
  0.950 &
  1.000 &
  -6.062 \\ \cmidrule(l){2-6} 
 &
  \multirow{2}{*}{} &
  \textbf{Disambigutation} &
  \multicolumn{2}{c}{\textbf{Content Fidelity}} &
  \textbf{Language Fluency} \\ 
 &
   &
  $\Delta \text{Acc}_\text{Am} \uparrow$ &
  $\Delta \text{Acc}_\text{Dis} \downarrow$ &
  Pathology Match $\uparrow$ &
  PLL  $\uparrow$ \\ \cmidrule(r){2-6} 
 &
  \textbf{KBR} &
  0.343 &
  0.001 &
  0.782 &
  -6.862 \\
 &
  \textbf{ST} &
  0.501 &
  0.051 &
  0.629 &
  -6.454 \\
 &
  \textbf{PPLM} &
  0.386 &
  0.115 &
  0.643 &
  -6.890 \\
 &
  \textbf{DEPEN} &
  0.500 &
  0.052 &
  0.676 &
  -6.529 \\
 &
  \textbf{Ours} &
  0.496 &
  0.032 &
  0.809 &
  -6.232 \\ \bottomrule\toprule
\multirow{9}{*}{\begin{tabular}[c]{@{}c@{}}\textbf{VA-}\\ \textbf{Annotated}\end{tabular}} &
   &
  \textbf{Ambiguity Acc.} &
  \textbf{Decision Acc.} &
  \textbf{Pathology Match} &
  \textbf{PLL} \\ 
 &
  \textbf{Raw Text} &
  0.955 &
   0.946 &
   1.000 & -5.652
   \\\cmidrule(l){2-6}
 &
  \multirow{2}{*}{} &
  \textbf{Disambigutation} &
  \multicolumn{2}{c}{\textbf{Content Fidelity}} &
  \textbf{Language Fluency} \\ 
 &
   &
  $\Delta \text{Acc}_{\text{Am}} \uparrow$ &
  $\Delta \text{Acc}_\text{Dis} \downarrow$ &
  Pathology Match $\uparrow$ &
  PLL  $\uparrow$ \\\cmidrule(r){2-6}
 &
  \textbf{KBR} &
   0.495 &
   0.007&
   0.885&
   6.109 \\
 &
 \textbf{ST} &
   0.311&
   0.235&
   0.351&
   -7.284\\
 &
  \textbf{PPLM} &
   0.270&
   0.146&
   0.582&
   -6.147\\
 &
  \textbf{DEPEN} &
   0.353&
   0.047&
   0.838&
   -6.102\\
 &
  \textbf{Ours} &
   0.481 &
   0.009&
   0.856&
   -5.821\\ \bottomrule
\end{tabular}%
}
\caption{Automatic Evaluation Results on OpenI- and VA-Annotated. Statistics about the original data is provided separately.}

\label{tab: automatic evaluation}
\end{table*}

The overall pipeline for building our annotated dataset is shown in \Cref{fig: data annotation}. We elaborate each dataset as follows. 
See more in Appendix Section ``Dataset Details''.   

\begin{figure}
     \centering
     \begin{subfigure}{0.493\linewidth}
         \centering
         \includegraphics[width=\linewidth]{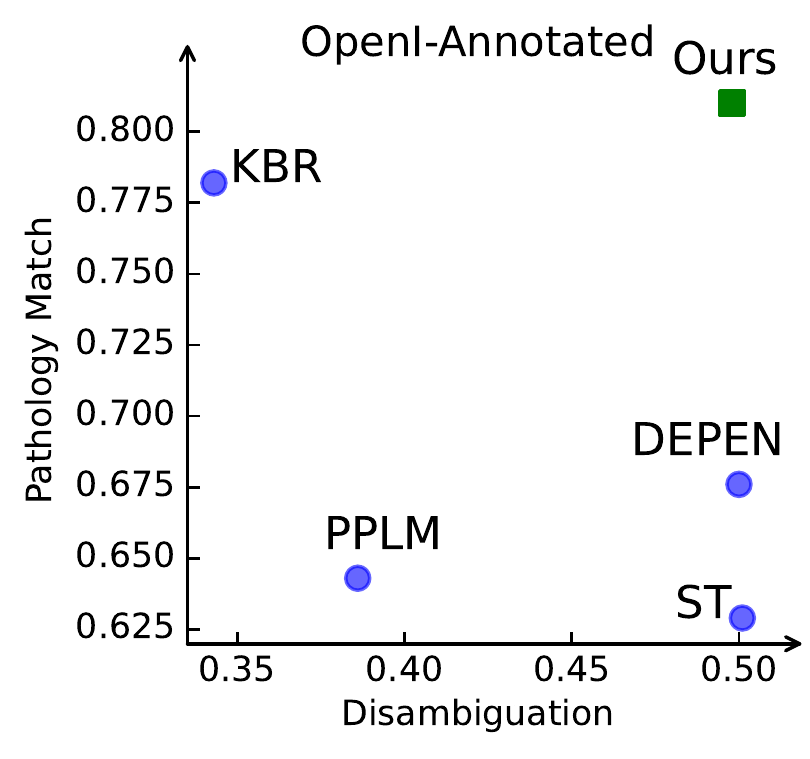}
         \vspace{-1.7em}
         \caption{}
         \label{fig: toxicity}
     \end{subfigure}
     \hfill
     \begin{subfigure}{0.493\linewidth}
         \centering
         \includegraphics[width=\linewidth]{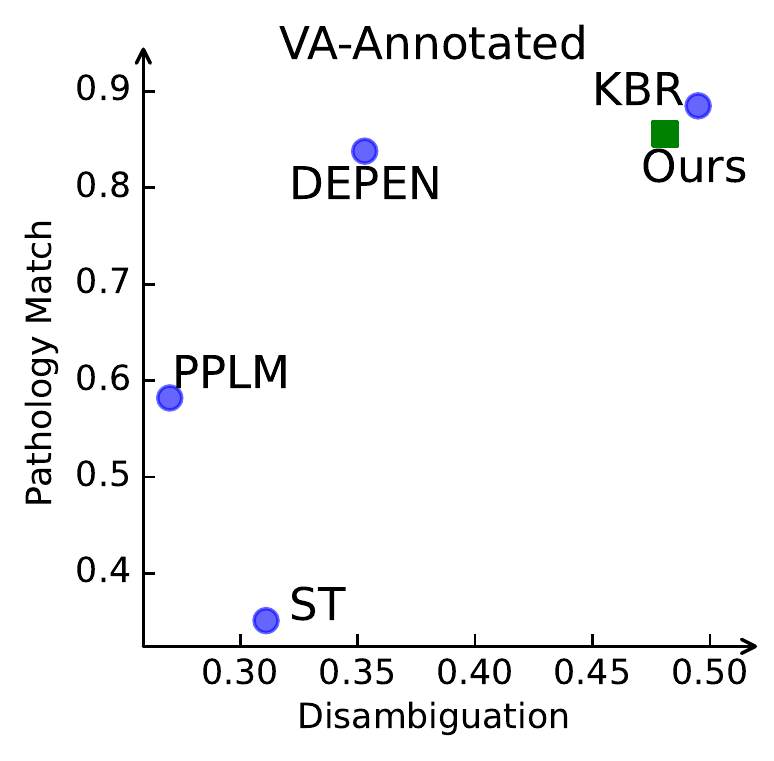} %need change to VA
         \vspace{-1.7em}
         \caption{}
         \label{fig: profession}
     \end{subfigure}
     \vspace{-0.5em}
     \caption{ Trade-off between Disambiguation and 
     %Fertility 
     Fidelity
     on (a) OpenI-Annotated (b) VA-Annotated. Higher disambiguation and pathology match (more upper-right corner in geometric) indicates a better rewriting. }
     \label{fig: disambiguation vs fedility}
     \vspace{-1em}
    \end{figure}

\paragraph{OpenI-Annotated} 

%We create OpenI-Annotated dataset and use it to rewrite in chest domain. OpenI is a large-scale high-quality dataset, hosted by IndianaNetwork for Patient Care \cite{demner2016preparing} and containing paired 7,470 frontal-view radiographs and radiology written reports. It is widely used in medical literature. To build our rewriting dataset, we take sentence-level labeled subset of OpenI released by~\citep{harzig2019addressing}.

%We create OpenI-Annotated dataset,  use it to rewrite in the chest domain. We first take the sentence-level subset corpus of OpenI, released and labeled by~\citep{harzig2019addressing}, where sentences are split from radiology written reports about chest.

We take the sentence-level subset of OpenI released by~\citet{harzig2019addressing}. 
% Our medical team 
Our medical team conducts data cleaning by removing identical sentences and completing missing terms (mistakenly masked in de-identification) according to their domain knowledge. We distinguish sentences that are irrelevant to our task and re-label sentences that contain abnormal findings and that are ambiguous according to corresponding criteria.  In the end, the OpenI-Annotated dataset consists of sentences with associated binary labels for being irrelevant, ambiguous, or abnormal. Statistics are shown in \cref{tab:datasets}. In the experiment, we split the OpenI-Annotated data to train/validation/test sets by 70\%, 10\%, 20\%.

\paragraph{VA Annotated}
We create the VA-Annotated dataset and use it in general-domain medical report rewriting. We use the VA radiology report corpus, recently introduced in~\citep{yan2022radbert}. As a general medical report corpus, it covers 8 modalities and 35 body parts for 70 modality-bodypart combinations.
We sample a subset and split them into sentences. 
Then similar data cleaning steps for OpenI-Annotated are used. Each sentence is annotated with binary labels for relevance, ambiguity and abnormality. We call it VA-Annotated and its statistics are listed in \cref{tab:datasets}. In the experiment, we split it into train/validation/test sets by 70\%, 10\%, 20\%.

\paragraph{Human Labeling Procedures}
In each experiment, experts start the labeling procedures after data cleaning.

First, for relevance labeling, the sentences only containing facts (e.g., \textit{CT 
%Images 
of the chest are taken}) and body parts (\textit{% This is the 
left knee was not evaluated}) are regarded as irrelevant to our task since there are no abnormal/normal diagnoses
% symptoms 
mentioned. Our medical team provide their binary labels for relevance.
Secondly, for relevant sentences, the medical team annotates a binary \textit{abnormality label}, indicating if there is an abnormal symptom found in the sentence. Sentences containing abnormal symptoms usually imply a diagnostic decision of being sick. Our non-expert team annotates an \textit{ambiguity label}, indicating whether the diagnostic decision looks too ambiguous for patients to understand.

The annotations are performed iteratively until inter-annotator Cohen's Kappa higher than substantial agreement ($\geq$ 0.8). The final discrepant labels were resolved by doctors in our medical team. 
See more details about the labeling criteria in Appendix Section ``Human Labeling Details''.

\subsection{Contrastive Pretraining Datasets}
Here we introduced the corpus used in contrastive pretraining and elaborate their fine-grained pathological labels about different health conditions.  

\paragraph{MIMIC-CXR} 
MIMIC-CXR is the largest public-domain chest x-ray dataset proposed in~\citep{johnson2019mimic} with 220k reports. We obtain the report sentences after de-duplication. 
For fine-grained pathological labels, we use CheXbert~\cite{irvin2019chexpert}, an automated deep-learning based chest radiology report labeler trained with MIMIC-CXR data (therefore no domain shift occurs), to label 14 fine-grained diseases. We keep sentences that have at most one disease noted. We end up with 1,129,478 sentences. The 14 diseases and statistics are listed in \cref{tab:mimic conditions}. This dataset is used in pretraining of the chest rewriting experiment. 

% \usepackage{graphicx}

\paragraph{VA-Rest}
The remaining unannotated sentences of the VA corpus~\cite{yan2022radbert} are used as a contrastive pretraining corpus. VA contains general medical reports covering different body parts,
%areas, 
therefore, CheXbert is not applicable.
%a capable labeler anymore. 
Instead, we use clustering results as pseudo-labels for different fine-grained pathological patterns.
We first obtain % get 
the sentence representations by feeding
% inputting 
them into a RadBERT model~\citep{yan2022radbert}, which is finetuned with the VA corpus by language modeling, and extracting the last hidden states. Then we reduce the dimension to $D$ with Uniform Manifold Approximation and Projection \cite{mcinnes2018umap}. Based on the reduced embeddings, sentences are clustered with a Gaussian Mixture Model into $K$ clusters. After experimenting with different parameters, we notice $K=14$ and $D=256$ achieve a good Silhouette score. 
See more in Appendix Section ``Clustering Details''.

\begin{table}[]
\resizebox{\columnwidth}{!}{%
\begin{tabular}{ccccc}
\toprule
\multicolumn{1}{l}{} & \multicolumn{2}{c}{\textbf{OpenI-Annotated}}                      & \multicolumn{2}{c}{\textbf{VA-Annotated}}       \\
Models               & Disam $\uparrow$       & Fidelity $\uparrow$             & Disam $\uparrow$ & Fidelity $\uparrow$ \\ \midrule
KBR       & 0.317 & 0.908 &  \textbf{0.488} & \textbf{0.988} \\
ST & 0.609    & 0.526 & 0.225  &0.350     \\
PPLM              & 0.376 & 0.624 & 0.333 & 0.575 \\
DEPEN             & 0.571 & 0.795 &  0.282		 & 0.941 \\
Ours                 & \textbf{0.792} & \textbf{0.921} &   0.383                        & 0.808                  \\ \bottomrule
\end{tabular}%
}
\caption{Human Evaluations. \textit{Disam}: Disambiguation.  }
\vspace{-1em}
\label{tab: human eval}
\end{table}

\begin{table*}[]
\centering
% \resizebox{0.9\textwidth}{!}{%
\normalsize
\begin{tabular}{ll}
\toprule
\multicolumn{2}{c}{\textbf{Contradictory Findings}}                                                  \\ \midrule
\textbf{Original Input}  &   \textit{normal} cardiac contour with \textit{atherosclerotic changes} throughout the aorta.                                                                       \\\cmidrule(l){1-2}
\textbf{KBR}   &      normal heart contour with atherosclerotic changes throughout the aorta.                                                                    \\
\textbf{ST}   &   normal cardiac contour with atherosclerotic changes throughout the aorta.                                                                       \\
\textbf{PPLM}             &      unchknown tortuous cardiac contour unchanged tortuous atherosclerotic changes throughout the aorta.                                                                   \\
\textbf{DEPEN}            &    diaphragmclerotic changes throughout the thoracic aorta. 
                                                                      \\
\textbf{Ours}             &      The cardiac contour shows atherosclerotic changes throughout the aorta.                                                                    \\ 
\bottomrule \toprule
\multicolumn{2}{c}{\textbf{Medical Jargon}}                                                          \\ \midrule
\textbf{Original Input}  &    
maybe \textit{secondary} to prominent mediastinal fat or tortuous.
% the mediastinal and hilar lymph nodes are less prominent than previously.	
                          \\\cmidrule(l){1-2}
\textbf{KBR}  &     
maybe secondary to prominent mediastinum palmitic acid or tortuous.
% the mediastinum and hilar lymph fluid nodes are smaller prominent than previously.                                    
\\
\textbf{ST}    &  

secondary to prior mediastinal or tortuous.                                       
\\
\textbf{PPLM}             &        optional secondary to the calcifiedsecondsmediastinal fat or tortuous. Include.
% more prominence on scarring is more prominent on  than on.                                
\\
\textbf{DEPEN}            &    
ouching compared the to  the mediastinal fat or tortuous.
% the mediastinal and hilar lymph nodes appear  less prominent as compared previously.                  
\\
\textbf{Ours}             &   
maybe due to the mediastinal fat or tortuous.
% the mediastinal and hilar lymph nodes are slightly  smaller than before. 
\\ 
\bottomrule
\end{tabular}%
% }
\caption{Examples of rewriting by different models for ambiguous sentences from OpenI-Annotated.}
% Writing examples of ambiguous sentences on OpenI-Annotated. This is a sentence containing abnormal symptoms.}
\vspace{-1.em}
\label{tab: case study example}
\end{table*}

\subsection{Baselines and Ablations}
\label{sec: baselines}
We follow the experiment design of \citet{xu2022self}, and choose baseline models that
are commonly used and have publicly available code:
\begin{itemize}
    \item \textbf{Knowledge-Based Replacement (KBR)} regenerates a sentence by replacing ambiguous terms with  unambiguous alternatives. Following the previous work \cite{qenam2017text}, we build a dictionary for replacement by looking up the Consumer Health Vocabulary.\footnote{Availalbe as a part of the UMLS \url{https://www.nlm.nih.gov/research/umls/index.html}
    %We will release our dictionary after publication to facilitate future research.
    } We notice that difficult special terms are also replaced with their layman language. %, which makes KBR a strong baseline.
    \item \textbf{Style Transformer~(ST)} A strong style-transfer model \cite{dai2019style} with adversarial training and a transformer architecture to transfer style while preserving content by reconstruction. 
    \item \textbf{Controllable Generation} We include two perturbation-based controllable generation models --   \emph{PPLM} \cite{dathathri2019plug} and \emph{DEPEN} \cite{he2021detect}. PPLM is a decoder-based language model but not capable of regeneration. In order to use it in our task, we modify it into a Seq2Seq model. We also adapt DEPEN so that it generates a less ambiguous sentence, as it is originally proposed for bias neutralization rewriting. 
\end{itemize}
%\textbf{Knowledge-based Replacement (KBR)} Regenerates a sentence by replacing ambiguous terms with its unambiguous alternatives. Following the previous work \cite{qenam2017text}, our medical team builds a dictionary for replacement by looking up the Consumer Health Vocabulary \footnote{We will release our dictionary after publishing to facilitate future research}. We notice that difficult special terms are also replaced with their layman language, which makes KBR is a strong baseline.
%\textbf{Style Transformer} A style transfer model \cite{dai2019style} which GAN-based idea to transfer style while preserving the content by reconstruction. 
%\textbf{Controllable Generation} We include two perturbation-based controllable generation models --   \emph{PPLM} \cite{dathathri2019plug} and \emph{DEPEN} \cite{he2021detect}. PPLM is a decoder-based language model and not capable for regeneration. In order to use it in our task, we modify it into Seq2Seq style. We also adapt DEPEN so that it generates a more tendentious sentence, as it is originally proposed for neutralization rewriting. 

After adapting PPLM and DEPEN, they can be regarded as ablations -- PPLM can be considered as
% is 
our algorithm without contrastive pretraining and `detect' steps, while DEPEN is ours without contrastive pretraining. 
 
\subsection{Evaluation Metrics}
\label{sec: automatic evaluation}
Following the evaluation of \citet{xu2022self}, we compare  rewritten results from the following aspects.
\begin{itemize}
    \item \textbf{Disambiguation:} We measure the level of ambiguity using the accuracy of a Bert classifier, which is finetuned to predict ambiguity labels in OpenI-Annotated or VA-Annotated. The accuracy deduction $\Delta \text{Acc}_\text{Am}$ is regraded as disambiguation performance.
    \item \textbf{Fidelity:} We evaluate fidelity at two granularities: (1) a coarse-grained one which evaluates the persistence of the original abnormality label, measured by the accuracy gap $\Delta \text{Acc}_\text{Dis}$ from a Bert classifier finetuned to predict abnormality.  (2) a fine-grained one which evaluates the 
    match
    of pathology, measured by the match rate of CheXbert labeled results or pseudo-labels.
    \item \textbf{Language Quality:}  Following \citet{he2021detect}, we use Pseudo-Log-Likelihood (PLL) \cite{salazar2020masked} score to measure language fluency.
\end{itemize}

\subsection{Human Evaluation} 
\label{sec: human evaluation}
Rewritten results generated with different models are reviewed by radiology experts. For an ambiguous sentence, the rewritten result and its associated abnormality labels are shown to reviewers simultaneously. Reviewers decide (1) if the rewriting is successful in disambiguation; and (2) if the original content has been preserved by rewriting. As for the second one, our reviewers have a rigorous objective that includes language quality evaluation -- a rewrite will be considered as a failure if there are any significant changes from the original findings or if proper English is not used.
We collect the results of human evaluations and calculate the disambiguation and fidelity success rates.  

\section{Results and Analysis}

%The automatic evaluation results of our model with baselines are shown in \cref{tab:OpenI automatic result} (OpenI-Annotated) and \cref{tab: VA automatic result} (VA-Annotated). In the task of disambiguation rewriting, we focus on both ambiguity mitigation and content fidelity. Therefore, we illustrate these two metrics in \cref{fig: disambiguation vs fedility}. We also report the human evaluation results on OpenI-Annotated and VA-Annotated in \cref{tab: human eval} and \cref{tab: human eval on VA}. 

\subsection{Performance Comparison}
The automatic evaluation results are shown in \cref{tab: automatic evaluation}. Notably, it is  sub-optimal to achieve the lowest ambiguity while generating a destroyed sentence. Therefore, we believe a good model is the one with an optimal balance between disambiguation rewriting and content preservation. We illustrate the trade-off between disambiguation and fidelity in \cref{fig: disambiguation vs fedility}, where the upper-right corner indicates a good model. Our rewriting model resides at the upper-right corner in the two experiments, indicating a superior balance between disambiguation rewriting and content fidelity. This also agrees with human evaluation results shown in \cref{tab: human eval}.

We discuss more about the results in the following. First, we compare our model with ST. Though it has a reasonable disambiguation (0.501 on OpenI-Annotated and 0.311 on VA-Annotated), ST has bad fidelity scores in both coarse-grained and fine-grained evaluations (the worst one on VA-Annotated). The generation quality is also worse compared with other models. We notice a rewriting from ST usually changes the original sentences significantly on both OpenI-Annotated and VA-Annotated, which explains why ST is able to disambiguate while fails in preserve fidelity. 
%disambiguate is fine while fidelity is bad. 
We provide our conjecture about the underlying reason: as an end-to-end model trained with multiple objectives at the same time,  ST is more fragile when balancing objectives, making it difficult to find the sweet point between rewriting for disambiguation and content preservation.

Then, we compare our model with controllable generation baselines -- PPLM and DEPEN. PPLM is a variation of our model without the detect step and constrastive pretraining step. Without the detect step, unnecessary edits can be  applied, as the model knows little about which parts are ambiguous. And without contrastive pretraining to inject distinguishable domain knowledge, the model will fail to preserve the main pathological content, having bad content fidelity in the end. Therefore, PPLM is not effective
% bad 
at both disambiguation and fidelity on both OpenI-Annotated and VA-Annotated. DEPEN shows improvements on disambiguation and maintains the original abnormality compared with PPLM, as the detect stage is added. But it fails to preserve fine-grained pathology match due to the lack of
% missing 
contrastive pretraining. Our model has the best overall performance in disambiguation and fidelity at different 
%grain-
%levels. 
granularities.
The improvement between PPLM, DEPEN, and ours indicates the effectiveness of each component in our model. 

We notice a clear difference in performance of KBR -- it fails to disambiguate in OpenI-Annotated while it becomes a strong baseline in VA-Annotated by achieving the best in both disambiguation and fidelity. 
We conjecture the reasons to be  domain difference. 
We discuss the divergence below. 

\subsection{Specific Domain vs.~General Domain}
%The different performance of the KBR model in OpenI-Annotated and VA-Annotated experiments raised our attention. We conjecture the underlying reason is due to the domain difference and discuss in the following. 

As one can observe in~\Cref{tab: automatic evaluation}, our neural rewriting model is able to substantially outperform other baselines on OpenI-Annotated~(specific domain). This indicates that given a reasonable amount of training data, our framework can perform well for a particular domain. On the VA dataset~(general domain), KBR becomes a strong baseline. We notice that when creating the dictionary, human medical experts are good at 
% %brainstorming 
proposing jargons across broadly different diseases and organs in general healthcare domains. However proposing terms that are specific to a domain requires deeper knowledge 
%nd longer working life 
in that particular discipline. Therefore, VA-Annotated is more well-covered by the dictionary than OpenI-Annotated which is specific to the chest domain. We found the dictionary coverage rate is 17.3\% on OpenI-Annotated while 21.5\% on VA-Annotated, which explains why replacing works better in VA-Annotated.
%We also calculate coverage rates on these two datasets (OpenI-Annotated: 17.3\%, VA-Annotated: 21.5\%) and the higher rate of VA-Annotated supports our conjecture. 

% %It 
% The different performance of KBR also suggests the benefits and drawbacks of KBR. When a comprehensive knowledge dictionary is available, replacement can be promising. Nevertheless, as it becomes more comprehensive, more human effort is required. Instead, deep learning methods alleviate human effort, while performance may be compromised in some situations.    

However, since knowledge-based models come with a price of dictionary compiling
% careful analysis 
%on the data 
and human (especially expert) effort,
%, especially those from experts, 
it may be difficult to extend them to solve domain-specific problems as each domain requires significant workload and expert experience. 
%Making context-dependent rewriting is difficult for them  
%from experts. 
Instead, our rewriting framework is potentially a more promising direction to explore for this task, % field, 
as it  alleviates human effort while achieving competitive or even better performance.

\vspace{-0.2em}

\subsection{Case Study}

% \textcolor{red}{@Chunnan: Please try to fill in examples  and write down the analysis here.}

% \textcolor{red}{Find at least 3 examples and fill in the table 8. For each example, write down:}

% \textcolor{red}{1. why the original text is ambiguous? ambiguous for what? why it makes patient confused?}

% \textcolor{red}{2. Interpretation about rewritten results from both baselines and our model. Why they are good and why they are bad.} 

We show some examples in \cref{tab: case study example}. 
More examples can be found in the Appendix Section ``Examples''.
Findings in the first example are labeled as abnormal. The contradictory usage of ``\textit{normal}'' and ``\textit{atherosclerotic changes}'' in the sentence makes patients confused about the abnormality. As shown in this example, KBR replaces the special term with layman language (\textit{cardiac} $\rightarrow$ \textit{heart}), but this does not help disambiguation since there is still a contradiction. These limitations suggest that replacement-based models cannot handle patterns outside the dictionary or patterns at the sentence level. ST fails to rewrite a sentence. PPLM suffers from repetition issues and generates output that is not comprehensible.  DEPEN can target the editing area with its detect step. However it fails to maintain fidelity without contrastive pretraining, and involves new findings
% symptoms 
that are inaccurate and change the original content drastically. However, our model achieves successful disambiguation by rewriting a contradiction-free sentence with minimal editing (\textit{normal} $\rightarrow$ \textit{The})  while maintaining fidelity by preserving the original abnormality 
% symptom 
(\textit{atherosclerotic changes}). 

%The second example is about medical jargon ``prominent'', which usually implies serious and obvious abnormality in the finding. However, in regular usage, it means something important and distinguished.  

Ambiguity in the second example is caused by  medical jargon  ``\textit{secondary to}'', which implies ``\textit{mediastinal or tortuous} '' is the reason for an abnormal finding.  However, in regular usage, it means % something 
``\textit{less important}'' which is not the case or ``\textit{coming after}'' which diminishes the causation. While other baselines either fail to disambiguate or introduce
% involves 
new content, ours is able to find a rewriting that mostly matches the context to describe the pathology causation. 

\section{Related Work}

\paragraph{AI for Medical Reporting}

Recent advances in AI have enabled novel applications involving medical reports. Medical report generation~\citep{li2018hybrid, chen2020generating, yan2021weakly} aims to automatically generate descriptions for clinical radiographs, which may alleviate the development workload of radiologists. 
Some recent works notice the communication gaps between medical professionals and patients, and focus on changing terminology in medical reports to lay-person terms, using replacement-based methods based on dictionary or lexical rules~\citep{qenam2017text, oh2016porter}, or leveraging style transfer techniques~\citep{xu2022self}. However, in this study, we highlight that the confusion also comes from ambiguity in readable texts where a knowledge base search may be insufficient. 

% \citet{mityul2018patient} considered three aspects that can be improved for patient-centered radiology reporting: report language, report format, and result delivery. AI/NLP report generators can help in report language. Report format and result delivery involve UI-UX design where AI may help, too, to deliver a total solution. In this paper, we focused on report language to train AI to deliver understandable and clinically accurate radiology reports. 

% With regard to improving report language, there are two aspects identified by~\citep{how2020reporting}: 1) Understandable report with little specialized language. 2) No ambiguity about the significance of findings.
% For the first aspect, we may replace all medical specialized terms with either lay-person understandable terms or a hyperlink to a document explaining the term to address the issue by either pattern-matching (see, e.g.,~\cite{kandula2010semantic,hogarth2017}) or more advanced deep learning for named entity recognition and normalization (e.g.,~\cite{Wright2019NormCoDD} for disease names). 

\paragraph{Controlled Text Generation}
The task of ccontrolling the output of a text generation model has been investigated recently. One line of research mainly focuses on the training process, by finetuning models with desired attributes~\citep{gururangan2020don} or creating a class-conditioned model~\citep{keskar2019ctrl}. Another direction is to design decoding approaches which are lightweight and effective. For example, 
\citet{yang2021fudge} train classifiers to predict whether an attribute can be satisfied. 
PPLM~\citep{dathathri2019plug} controlls generation by updating a pretrained model's hidden states, and DEPEN~\cite{he2021detect} adapted PPLM into the seq2seq model to rewrite a neutral output.
There is less consideration of content preservation and domain knowledge in previous works, limiting their potential in sensitive or rigorous disciplines like healthcare. In contrast, our approach incorporates a contrastive pretraining step that effectively infuses domain knowledge and enhances content fidelity.  

\paragraph{Contrastive Learning}
Contrastive learning~\citep{gutmann2010noise, oord2018representation} learns representations by pulling positives together and negatives far from each other. Recent work has explored contrastive learning for NLP applications, such as contrastive data augmentation~\citep{shen2020simple, qu2020coda}, sentence embedding learning~\citep{kim2021self} and text generation generation~\citep{yan2022personalized, zhu2022visualize}.
Our work differs in that we apply a label-aware contrastive learning method to the pretraining stage of a language decoder, and mainly focus on its application to the healthcare domain.

\section{Conclusion}
Sharing medical information, especially reports with patients is essential to patient-centered care. Due to the communication gap between audiences, there is always ambiguity in reports, leading patients to be confused about their exam results. We collect and annotate two datasets containing radiology reports from healthcare systems in this study. We analyze and summarize three major causes of ambiguous reports: jargon, contradictions, and misleading grammatical errors, and propose a framework for disambiguation rewriting. Experimental results show that our model can achieve effective disambiguation while maintaining content fidelity.   

%Rather than resolving a challenging real-world problem in one work with the best performance, we aim to provide related analysis and raise future attention. 
Medical reporting is time-consuming and labor-intensive. Our work aims to inspire more research so that in the future, more AI systems like ours can be created and used to assist real-world medical services under expert auditing.  

\section{Ethics Statement}
The use of medical report data from VA received ethical approval from the IRB approval ID \#H200086. All patient information was de-identified and patient consent was waived. 
Despite the intention to mitigate ambiguity, an AI system may have unexpected outputs. Also rewriting a sentence may subtly alter its detailed content. A malicious user could adversarially uses the unexpected results. Hence, we emphasize that a system like ours should best be used with expert auditing to ensure safety. Reporting ambiguity is a practical task. Our work aims to establish a benchmark for this new problem and present a possible solution using NLP techniques to alleviate human effort. As an initial attempt, we hope to inspire more fut
ure research on this challenging problem.

\section{Acknowledgments}
This work was financially supported in part by the Office of the Assistant Secretary of Defense for Health Affairs through the Accelerating Innovation in Military Medicine (AIMM) Research Award program W81XWH-20-1-0693 and in part by the National Science Foundation Award \#1750063. Opinions, interpretations, conclusions and recommendations are
those of the author and are not necessarily endorsed by the funding agencies. Zexue He is funded by IBM Ph.D. Fellowship. 

\bibliography{aaai23}

\clearpage
\appendix

\section{Human Labeling Details}
Our medical team proposes the following guidelines for the concept : \textit{abnormality}, \textit{irrelevance}, and \textit{ambiguity}.
\subsection{Criteria for Irrelevance}
\label{sec: criteria}
Sentences only containing following information are regarded as irrelevant:
\begin{itemize}
    \item How imaging is taken (e.g., PA and lateral views are obtained).
    \item A body part and nothing else. (eg., chest, knee).
    \item Communication. (e.g., ordering doctors are informed).
    \item Unfinished sentences (sentence splitting errors).
    % \item Too many errors about unknown words that are  unlikely to be any PHI that need to be de-id’ed) 
\end{itemize}

\subsection{Criteria for Abnormality}
Sentences in the following situations are considered to have diagnosis as abnormal:
\begin{itemize}
    \item Multiple copies of the same example sentences (i.e., not ``unique'') if labeled differently, will be relabeled to ensure all of them have the same correct label. (These examples may be considered  ``confusing'' if not all of them are ambiguous. Some inconsistent labels are just human errors).
    \item Sentences suggesting further or followup exams, including  ``document resolution'', are abnormal.
    \item Reporting limitations of exams only, including patient rotation, inspiration, and inspiratory effort, are abnormal, unless normal or no findings clearly stated.
    \item Foreign objects, artifacts (e.g., picc, catheter, wire, tube, stent, pacemaker, etc), past surgical marks (e.g., cabg, sternotomy, post-op) are abnormal, malpositioned or well positioned regardless. 
    \item Shadow of body parts, especially nipples, nipple piercings, breast mass, breast implant, and trachea, are abnormal.
    \item Healed and resolved prior conditions are normal.
    \item Improved, including significantly improved and minimal conditions are abnormal.
    \item Stable and unchanged conditions, if from a prior abnormal condition, are abnormal. Likewise, if from a normal condition, they are normal. 
    \item However, stable or unchanged body parts, if no condition is given, are abnormal. These are actually context dependent.
\end{itemize}

\subsection{Criteria for Ambiguity}
Sentences in the following situations are considered as ambiguous sentences:
\begin{itemize}
    \item Jargon: Daily words with special meanings in radiology reports, such as \textit{unremarkable}, \textit{nonspecific}, \textit{prominent}, etc. 
    \item Stable, unchanged body parts where abnormality is context dependent.
   \item Contradicting decisions in the same sentences. 
   \item Misleading Grammar errors, e.g., no period between multiple sentences.
   \item Uncertainty clearly stated is not ambiguous.
\end{itemize}

\section{Discussion about Ambiguity Definition}
The medical team define ``ambiguity'' and create an annotation guideline based on review of the literature~\cite{how2020reporting,mityul2018patient,gunn2013quality}, feedback from patients, and online resources including blogs\footnote{https://radiologyinplainenglish.com/}$^,$\footnote{https://radiopaedia.org/Radiopaedia}, patient forums and social media.
%(e.g., Many questions about ``unremarkable'' in medical reports in Quora  etc.) 

Here we add a comparison discussion. (1) \textit{uncertainty} vs. \textit{contradictory}:  an uncertain statement is not ``ambiguous'', as it is not introduced by the ambiguous writing of report writers (the medical professionals are uncertain about the diagnosis results as well), whereas a contradicting statement is ambiguous. (2) medical jargon vs. terminology terms: medical jargon (\eg, “unremarkable” and “interval appearance of \ldots”) are considered ambiguous as their usage is different from general case, while terminology terms (\eg, “granuloma” and “plural effusion”) are not because they have already been clearly defined in the medical literature. See Section Criteria for Ambiguity for details.
% % \noindent\textbf{References}

% % A1. Alarifi, M., Patrick, T., Jabour, A., Wu, M., & Luo, J. (2021). Understanding patient needs and gaps in radiology reports through online discussion forum analysis. Insights into imaging, 12(1), 1-9.

% % A2. \url{https://radiologyinplainenglish.com/} Radiology in plain English.

% % A3. \url{https://radiopaedia.org/} Radiopaedia.

\section{Dataset Details}
\subsection{OpenI-Annotated Dataset}
OpenI is a large-scale high-quality dataset, hosted by Indiana Network for Patient Care \cite{demner2016preparing} and containing paired 7,470 frontal-view radiographs and radiology written reports. \citep{harzig2019addressing} takes a subset of it,  split reports into sentences, and labeled each sentence with a label for abnormal findings.

The sentences in \citep{harzig2019addressing} are anonymized with a de-identification software package, which is overly sensitive to ensure that no patient private information will be released. However, it results in a large number of incomprehensible sentences as important terms are masked mistakenly by de-identification. Our medical team conduct data cleaning by removing identical sentences and completing the masked term with their domain knowledge. We also noticed that the original labels in \citep{harzig2019addressing} contains inconsistency. Therefore, our medical team re-label the abnormality of the sentences % that contains abnormal findings 
according to the criteria outlined 
% introduced 
in Section Criteria for Abnormality.  

\subsection{VA-Annotated Dataset}
VA radiology report corpus%, 
% introduced  in~\citep{yan2022radbert}, 
 is a general domain corpus includes 150 million radiology reports from 130 VA facilities
nationwide during the past 20+ years. As a general medical report corpus, it covers 8 modalities, 35 body parts and 70 modality-body part combinations.
We sample a subset of VA dataset and split it into sentences. 
Our medical team conduct  data cleaning similar to OpenI-Annotated. Then, each sentence is annotated with binary labels for relevance, ambiguity and abnormality. 

\section{Implementation Details}
\subsection{Models and Parameters}
All transformers are implemented based on the  HuggingFace libraries \footnote{\url{https://huggingface.co/}}. 

\paragraph{Seq2Seq Model}
The Seq2Seq model we used in contrastive prertaining and rewriting is BART\cite{lewis2020bart}. We load weights of a pretrained BART (\texttt{distilbart-cnn-12-6}) from HuggingFace, with total parameters 306M. 
\paragraph{Detect Model} The model in the detect step is a finetuned \texttt{bert-base-uncased} which has 110M parameters. The attention score of each token with respect to \texttt{[CLS]} in the last layer is used as a salient score to measure the predictability for being ambiguous. 
\paragraph{Classifiers} 
Evaluation classifiers for abnormality, ambiguity and fine-grained disease are finetuned \texttt{bert-base-uncased} on annotated labels, with 110M parameters. 
\paragraph{Tokenizer} 
We use \texttt{nltk.tokenize.sent\_tokenize}\footnote{https://www.nltk.org/api/nltk.translate.html} from \texttt{nltk} library to split sentences in VA report corpus. 
\paragraph{CheXBert}:
We use official library\footnote{https://github.com/stanfordmlgroup/CheXbert} of CheXBert and loaded the released checkpoint pretrained on OpenI Corpus.   

\paragraph{RadBert}
We use the released pretrained model, specifically RoBERTa-4M\footnote{https://github.com/zzxslp/RadBERT} to extract sentence embeddings and generate clustering labels for VA data.

\subsection{Hyperparameters and Experiment Environment}
We list the hyperparameters used in our experiments in \cref{tab:hyperparameters}. ach experiment is repeated 3 times with different random seeds. All codes are implemented with Python3.8 and PyTorch1.7.1 with CUDA10.1. E
% \usepackage{graphicx}
\begin{table}[]
\centering
\resizebox{\columnwidth}{!}{%
\begin{tabular}{cc|cc|cc}
\toprule
\multicolumn{2}{c|}{\begin{tabular}[c]{@{}c@{}}Classifier \\ Finetuning\end{tabular}} &
  \multicolumn{2}{c|}{\begin{tabular}[c]{@{}c@{}}Contrastive \\ Pretraining\end{tabular}} &
  \multicolumn{2}{c}{\begin{tabular}[c]{@{}c@{}}Perturbation-based\\ Rewriting\end{tabular}} \\ \midrule
lr &
  1e-4 &
  lr &
  5e-4 &
  Iteration Times &
  15 \\
Batch Size &
  64 &
  Batch Size &
  256 &
  $\gamma$ &
  0.5 \\
Optimizer &
  Adam &
  Max Length &
  50 &
  Step Size &
  0.5 \\
Weight Decay &
  1e-5 &
  Temperature $\tau$  &
  0.07 &
  KL-loss Coef. &
  0.01 \\
Max Length &
  100 &
  $\lambda_1$ &
  1.0 &
  Max Length &
  50 \\
  Epochs&
  10 &
  $\lambda_2$ &
  1.0 &
 $\gamma$-scaling  Term &
  0.98 \\ \bottomrule
\end{tabular}%
}
\caption{Hyperparameters in experiments. \textit{lr} is shorten for \textit{learning rate}, \textit{coef.} is shorten for \textit{coefficient}.}
\label{tab:hyperparameters}
\end{table}

Our contrastive pretraining is operated on a Ubuntu (18.04.6 LTS) server with 4 % of 
NVIDIA Tesla \textregistered V100 GPUs. Each of them has 32GB  memory.  Our classifier finetuning and perturbation-based rewriting is operated on a Ubuntu (16.04.7 LTS) server with 4 NVIDIA GeForce GTS \textregistered 1080Ti GPUs. Each has memory of 11GB. 

\subsection{Clustering Details}
\begin{figure}
    \centering
    \includegraphics[width=\linewidth]{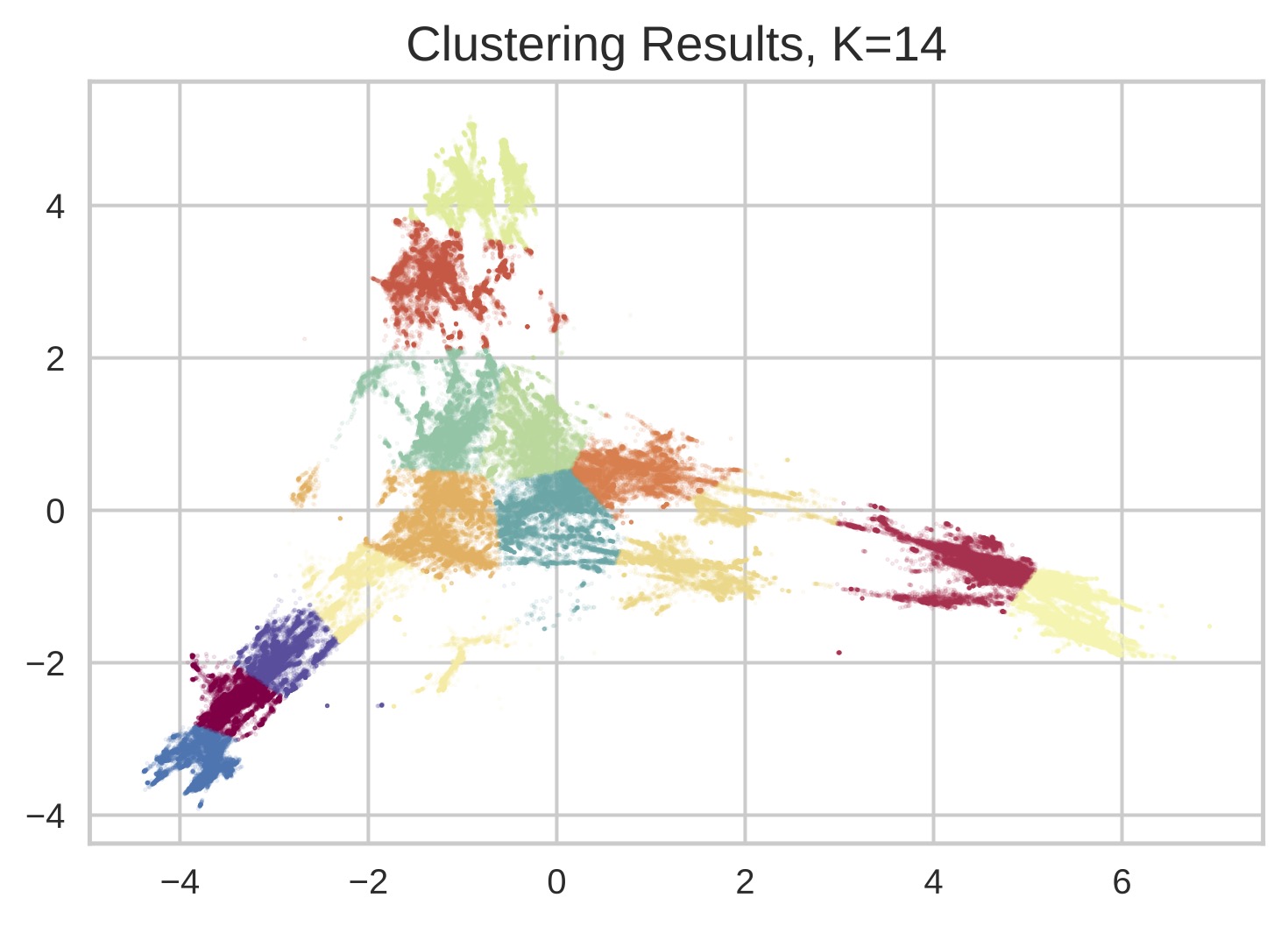}
    \caption{\small Clustering Visualization.  }
     \label{fig: cluster}
     \vspace{-1em}
    \end{figure}

When fine-grained disease labels are not available, we use clustering-based method to get pseudo-labels. We use official repository of \texttt{umap.UMAP} to first reduce the dimension to 256, then use \texttt{sklearn.mixture.GaussianMixture} (GMM) to cluster them into 14 clusters. We set n\_components=14, n\_init=3, random\_state=42 in GMM, and initialized the weights with KMeans. This setting enables us with the best silhouette score (0.50). The 2-dimension clustering visualization is shown in \cref{fig: cluster} after principal component analysis (PCA). 

\subsection{Evaluation Details}

\paragraph{Automatic Evaluation} For classification, we calculate accuracy scores with \texttt{scikit-learn(1.1.2)} library. For generation quality, We use the official repository \footnote{\url{https://github.com/awslabs/mlm-scoring.git}} to calculate the Pseudo-Log-Likelihood scores of generated sentences. To evaluate clustering performance, we use \texttt{sklearn.metrics.silhouette\_score} to calculate the silhouette score based on Euclidean Distance. 

\paragraph{Human Evaluation}
The evaluation judges the quality of a rewriting by the two criteria shown in the following.

First, the disambiguation -- does the rewriting successfully disambiguate the original sentence? The judgment follows the same criteria as the annotation guideline of ambiguity regardless of how it compares to the original sentence. That is, if the rewriting results in a sentence that is not ambiguous then it is labeled ``succeed'' and ``fail'' otherwise.

Second, the fidelity -- does the rewriting preserve medical contents of the original sentence? More precisely, if the original sentence states that an \emph{abnormality} is observed in a \emph{body part} with levels of \emph{severity}, \emph{acuteness} and \emph{certainty}, the rewriting must match the type of \emph{abnormality}, \emph{body part}, \emph{severity}, \emph{acuteness} and \emph{certainty} to be labeled as a ``succeed'' otherwise it is a ``fail.'' For example, if the original sentence is ``\ldots lungs are clear \ldots'' but the rewriting is ``\ldots left lungs are clear \ldots'' then it is a ``fail'' because of the extra ``left'' introduced. Moreover, if the rewriting contains serious grammar errors though the rewriting matches all elements, it will still be a ``fail.''

\section{Examples}

\subsection{Examples of Knowledge-based Dictionary}
As shown in \cref{tab:KBR dictionary}, we provide some examples about the dictionary used in knowledge-based replacement baseline (KBR). The dictionary is created by our medical team looking up from the Consumer Health Vocabulary (CHV). 
% We notice that the 
The dictionary not only includes medical jargon (e.g., \textit{collapse} $\rightarrow$ \textit{physiological shock}), but also involves difficult terminologies (e.g., \textit{pulmonary} $\rightarrow$ \textit{lung}) and abbreviations (e.g., \textit{picc line}), % which composes to 
providing a comprehensive medical dictionary covering different domains. 
% \usepackage{graphicx}
% \usepackage{graphicx}
\begin{table}[]
\centering
\resizebox{\columnwidth}{!}{%
\begin{tabular}{l}
\toprule
\multicolumn{1}{c}{Examples}                                  \\
\midrule
pulmonary edema $\rightarrow$ lung edema                      \\
cardiomegaly $\rightarrow$  enlarged heart                      \\
trachea $\rightarrow$  windpipe                                 \\
bony abnormalities $\rightarrow$  deformity of bone             \\
picc line $\rightarrow$  peripherally inserted central catheter \\
collapse $\rightarrow$  physiological shock                     \\
tenderness $\rightarrow$  sore to touch                         \\
thorax $\rightarrow$  chest                                     \\
tomogram $\rightarrow$  tomography                              \\
ectasia $\rightarrow$  abnormal dilation\\
pneumothoraces $\rightarrow$ free air in the chest outside the lung\\ 
calcifications $\rightarrow$ calcium deposit
\\aortic atherosclerotic vascular calcification $\rightarrow$ aortic calcification \\

\bottomrule
\end{tabular}%
}
\caption{Example items in the %created 
dictionary used in KBR. }
\label{tab:KBR dictionary}
\end{table}

\subsection{Rewriting Examples}

More rewriting examples are shown in \cref{tab: more study example}. The ambiguity in the first example is caused by a grammatical error ``sene''. KBR, PPLM and DEPEN can't fix this typo, and ST generates a new sentence with a content change. Our model can locate the area that causes ambiguity and correct it according to its surrounding context and infused domain knowledge.

The ambiguity in the second example is about medical jargon ``patent'', which usually refers to be ``obvious'' when used as an adjective word. However, in the medical domain, it means an object is not blocked by other parts, which emphasizes that no objects are obstructing it. Only our model can demonstrate that ``patent'' here amplifies the unobstructness of the interested tissues.

The ambiguity in the third example is caused by the contradictory writing -- ``not $<$\textit{an abnormal finding}$>$ which has $<$\textit{abnormal finding}$>$''. DEPEN generates a proper sentence, however, it involves yet another disease ( ``interstitial lung disease''). While KBR or ST fails to rewrite the sentence and PPLM outputs a destroyed sentence, our model not only fixes the contraction but also fixes the jargon ``prominence'' with ``abnormality'' too. 

The last example is about medical jargon ``prominence'', which implies serious and obvious abnormality in the finding. However, in regular usage, it means something important and distinguished. It has an opposite connotation between its regular usage and in medical reports. While other baselines either fail to rewrite or change the diagnosis dramatically, ours is able to find a rewriting that matched the context to describe lymph nodes.

\begin{table*}[]
\centering
\resizebox{\textwidth}{!}{%
\begin{tabular}{ll}
\toprule
% \multicolumn{2}{c}{\textbf{Misleading Grammatical Error}}                                            \\ \midrule
% Original Input  & \multicolumn{1}{l}{status post resection of left upper lobe.}           \\\midrule
% KBR  & status post surgical removal in left superior lobe                       \\
% ST   & status post surgical removal in left                                 \\
% PPLM             & status post resection of left upper lobe lobe                            \\
% DEPEN            & infiltrates status post opacifications of left upper lobe.Advertisements \\
% Ours             & Stable postop changes of left upper lobe                                 \\ \bottomrule
\multicolumn{2}{c}{\textbf{Misleading Grammatical Error (VA-Annotated)}}                                            \\ \midrule
\textbf{Original Input}  & \multicolumn{1}{l}{patchy changes are \colorbox{pink}{sene} bilateral.	
}           \\\midrule
\textbf{KBR}  & patchy changes are sene bilateral.                      \\
\textbf{ST}   & grossly, levels, bilateral .                                 \\
\textbf{PPLM}             & Fibrotic changes are sene bilateral.	                       \\
\textbf{DEPEN}            & these changes are sene bilateral. \\
\textbf{Ours}             & the changes are seen bilateral.                       \\ \bottomrule \toprule
\multicolumn{2}{c}{\textbf{Medical Jargon (VA-Annotated)}}                                                          \\ \midrule
\textbf{Original Input}  &    both neural foramina are \colorbox{pink}{patent}.
                                                                      \\\midrule
\textbf{KBR}  &     both neural foramina are patent.	                                                                     \\
\textbf{ST}    &      both neural foramina are patent .                                                               \\
\textbf{PPLM}             &       both neural foramina are patent.                                                                      \\
\textbf{DEPEN}            &   both neural foramina are patent.                                          \\
\textbf{Ours}             &    both neural foramina are open.                                                                       \\ \bottomrule
\toprule
\multicolumn{2}{c}{\textbf{Contradictory Findings (OpenI-Annotated)}}                                                  \\ \midrule
\textbf{Original Input}  &   is \colorbox{pink}{not} diffuse interstitial prominence which \colorbox{pink}{has} chronic appearance.                                                                       \\\midrule
\textbf{KBR}   &      is not diffuse interstitial prominent which has chronic appearance.                                                                    \\
\textbf{ST}   &   is not diffuse interstitial markings which may represent appearance .
                                                                       \\
\textbf{PPLM}             &      isscar formationalsooronopostopoposteroph nodopod nodogoduboidubzyoc lingophoc cy tuboidubod                                                                     \\
\textbf{DEPEN}            &    is also chronic diffuse interstitial lung disease which has chronic appearance.
                                                                      \\
\textbf{Ours}             &      is a  diffuse interstitial abnormality which has chronic appearance.                                                                    \\ \bottomrule \toprule
\multicolumn{2}{c}{\textbf{Medical Jargon (OpenI-Annotated)}}                                                          \\ \midrule
\textbf{Original Input}  &    the mediastinal and hilar lymph nodes are \colorbox{pink}{less prominent} than previously.
                                                                      \\\midrule
\textbf{KBR}  &     the mediastinum and hilar lymph fluid nodes are smaller prominent than previously.                                                                     \\
\textbf{ST}    &      the mediastinal and hilar lymph nodes are clear.                                                                    \\
\textbf{PPLM}             &       the mediastinal and hilar lymph nodes appear  less prominent as compared previously.                                                                                    \\
\textbf{DEPEN}            &   more prominence on scarring is more prominent on  than on.                                           \\
Ours             &   the mediastinal and hilar lymph nodes are slightly  smaller than before.                                                                       \\ \bottomrule
\end{tabular}%
}
\caption{Rewriting examples of ambiguous sentences in VA-Annotated (the top two) and OpenI-Annotated (the bottom two).}
\label{tab: more study example}
\end{table*}

\end{document}